\documentclass{article}
\usepackage{spconf,amsmath,graphicx}
\usepackage{tabularx} 
\usepackage{booktabs}
\usepackage{amssymb}
\usepackage[table]{xcolor}
\usepackage{colortbl}      % 让 \cellcolor 能用于表格单元
\usepackage{caption}
\usepackage{enumitem}
\usepackage[colorlinks=true, linkcolor=blue, citecolor=blue, urlcolor=blue]{hyperref}
\title{ICPO: Illocution-Calibrated Policy Optimization for Multi-Turn Conversation}
\name{\small Zhebo Wang$^{1}$~~Xiaohu Mu$^{3}$~~Zijie Zhou$^{4}$~~Mohan Li$^{5}$~~Wenpeng Xing$^{1,2}$~~Dezhang Kong$^{1,2,\dagger}$~~Meng Han$^{1,2,\dagger}$}
\address{\small
$^{1}$ Zhejiang University \quad
$^{2}$ Binjiang Institute of Zhejiang University \\
\small $^{3}$ University of Illinois Urbana-Champaign \quad
$^{4}$ China University of Petroleum (Beijing) \quad
\small $^{5}$ Guangzhou University \\
\small $^{\dagger}$ Corresponding authors: kdz@zju.edu.cn, mhan@zju.edu.cn
}
\begin{document}
\ninept
\maketitle
\begin{abstract}
Large Language Models (LLMs) in multi-turn conversations often suffer from a ``lost-in-conversation'' phenomenon, where they struggle to recover from early incorrect assumptions, particularly when users provide ambiguous initial instructions. We find that standard post-training techniques like Reinforcement Learning with Verifiable Rewards (RLVR) exacerbate this issue by rewarding confident, direct answers, thereby inducing overconfidence and discouraging the model from seeking clarification. To address this, we propose Illocution-Calibrated Policy Optimization (ICPO), a novel training framework that sensitizes the model to instruction ambiguity. ICPO augments the training corpus with underspecified prompts and conditions the reward signal on the user's illocutionary intent, rewarding the model for expressing uncertainty or asking for clarification when faced with ambiguity. Experiments demonstrate that ICPO fosters appropriate humility, yielding a substantial average improvement of 75\% in multi-turn conversation, while preserving robust performance on single-turn benchmarks. Our work presents a practical path toward more robust and collaborative conversational AI that can better navigate the nuances of human interaction.
\end{abstract}

\begingroup
\renewcommand\thefootnote{}
% \footnotetext{\,\textsuperscript{*}Equal contribution. \textsuperscript{\dag}Corresponding author.}
\footnotetext{©~2026 IEEE. Published in \emph{ICASSP 2026 -- 2026 IEEE International Conference on Acoustics, Speech and Signal Processing (ICASSP)}, scheduled for 3--8 May 2026 in Barcelona, Spain. Personal use of this material is permitted. However, permission to reprint/republish this material for advertising or promotional purposes or for creating new collective works for resale or redistribution to servers or lists, or to reuse any copyrighted component of this work in other works, must be obtained from the IEEE. Contact: Manager, Copyrights and Permissions / IEEE Service Center / 445 Hoes Lane / P.O. Box 1331 / Piscataway, NJ 08855-1331, USA. Telephone: +1~908~562~3966.}
\endgroup

\setcounter{footnote}{0}
\renewcommand\thefootnote{\arabic{footnote}}

\section{Introduction}
\label{sec:1}

Today's Large Language Models (LLMs) primarily interact with users through multi-turn conversation, such as in graphical chat interfaces and Agent-based applications~\cite{kong2025survey, xu2025copyrightprotectionlargelanguage, r1, r2, r3, r4, r5, r6, r7, r8, r9, r10}. In practice, users often begin with ambiguous or underspecified instructions and gradually clarify their needs through iterative exchanges—a process that is susceptible to the model's conversational priors~\cite{herlihy2024overcomingmiscalibratedconversationalpriors}. Recent studies report a pronounced ``lost-in-conversation'' phenomenon in multi-turn settings: once the model adopts an incorrect assumption or produces an inappropriate response early on, it often struggles to recover, exhibiting degraded performance and increased instability~\cite{laban2025llmslostmultiturnconversation}.

Unfortunately, to our knowledge, there have not been studies aiming to address the ``lost-in-conversation'' problem. Existing techniques like Reinforcement Learning (RL)~\cite{zhang2025surveyreinforcementlearninglarge, liu2025understandingr1zeroliketrainingcritical, yu2025dapoopensourcellmreinforcement, dong2025rlpluscounteringcapabilityboundary} can only improve single-turn performance, lacking the ability to handle multi-turn conversations. For example, it can induce overconfidence and mode collapse by compressing policy entropy, as the model concentrates probability mass on a few high-reward responses to maximize rewards. This reduces response diversity, impairs exploration, and ultimately results in rigid, unchanging response patterns~\cite{cui2025entropymechanismreinforcementlearning}. Besides, RL diminishes the tendency to seek clarification or express uncertainty in multi-turn conversations. This exacerbates the tendency to get ``lost'': once the trajectory deviates, low-entropy, high-stickiness strategies accumulate over multi-turn conversations, rendering the model unable in subsequent turns to overturn prior assumptions and leading to excessive reliance on previously erroneous strategies.

To address these limitations, we argue that multi-turn optimization must be sensitive to the illocution of the user's instruction, rather than treating all turns as requests for definitive solutions. Specifically, many degradation phenomena observed in multi-turn conversation~\cite{laban2025llmslostmultiturnconversation} stem from the mismatch between users' ambiguous instructions and low-entropy strategies. As a result, if a model knows when to clarify, question, or negotiate, it can preserve exploratory space to avoid hasty conclusions and improve recoverability.

Based on the above analysis, we propose ICPO: Illocution-Calibrated Policy Optimization. ICPO introduces an ambiguity-aware training loop that (i) augments the training corpus with underspecified prompts representative of real-world usage, and (ii) conditions the reinforcement signal on the illocutionary status. ICPO is a plug-in, model-agnostic addition to Reinforcement Learning with Verifiable Rewards (RLVR), applicable to various downstream tasks. Experiments show that ICPO promotes calibrated humility, a trait manifested as sustained high entropy and greater generative diversity. This yields a substantial average improvement of 75\% in multi-turn conversation, while preserving robust performance on single-turn benchmarks. Our key contributions in this paper are as follows.
\begin{itemize}[leftmargin=*, nolistsep]
    \item We first uncover a unique phenomenon that models trained via standard RLVR tend to become overconfident: when faced with ambiguous or under-specified instructions, they default to confident responses, exacerbating the tendency to get ``lost'' and degrading multi-turn performance.
    \item We propose ICPO, a novel framework that dynamically modulates the reward signal based on the illocution of the user's instruction, thereby maintaining humility during the RLVR process.
    \item Compared to standard RLVR approaches, ICPO improves reasoning ability significantly while maintaining appropriate humility and achieves a 75\% enhancement in multi-turn conversation performance. 
\end{itemize}

\section{RLVR Impairs Multi-Turn Conversation}
\label{sec:2}

\begin{figure}
    \centering
    \includegraphics[width=0.85\linewidth]{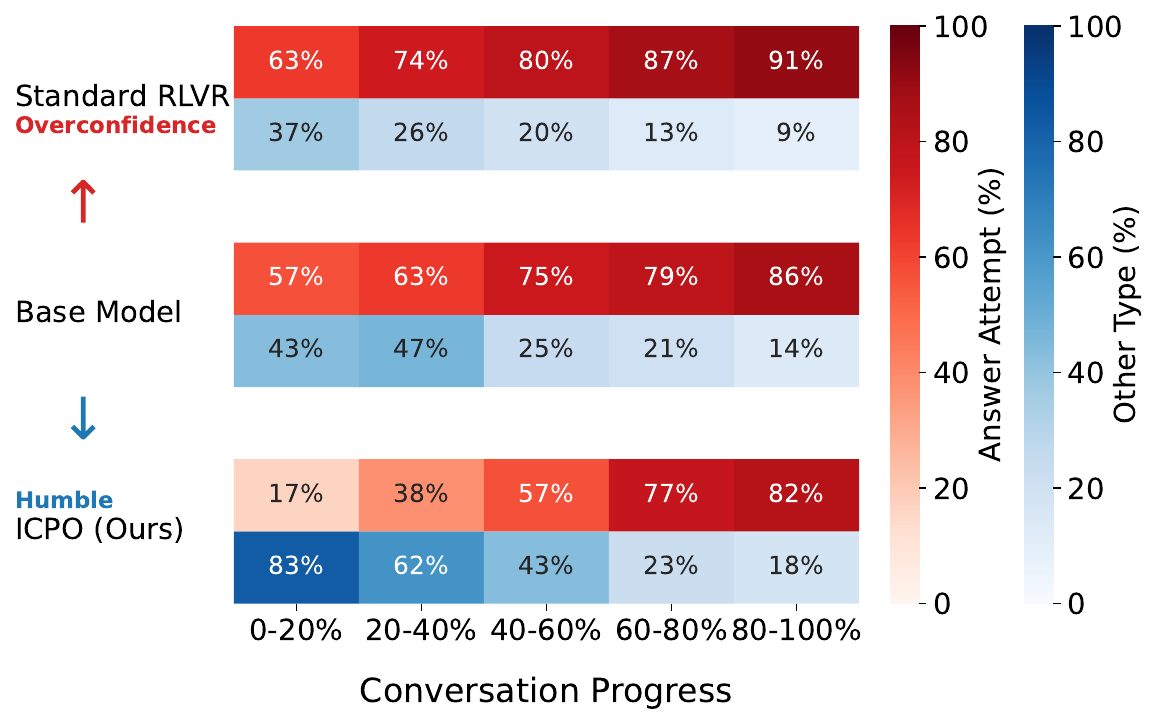}
    \caption{Average distribution of response type during multi-turn conversations starting with ambiguous instructions, in which the instruction conditions gradually become complete as the conversation turns progress.}
    \label{fig:f1}
\end{figure}

\begin{figure}
    \centering
    \includegraphics[width=0.87\linewidth]{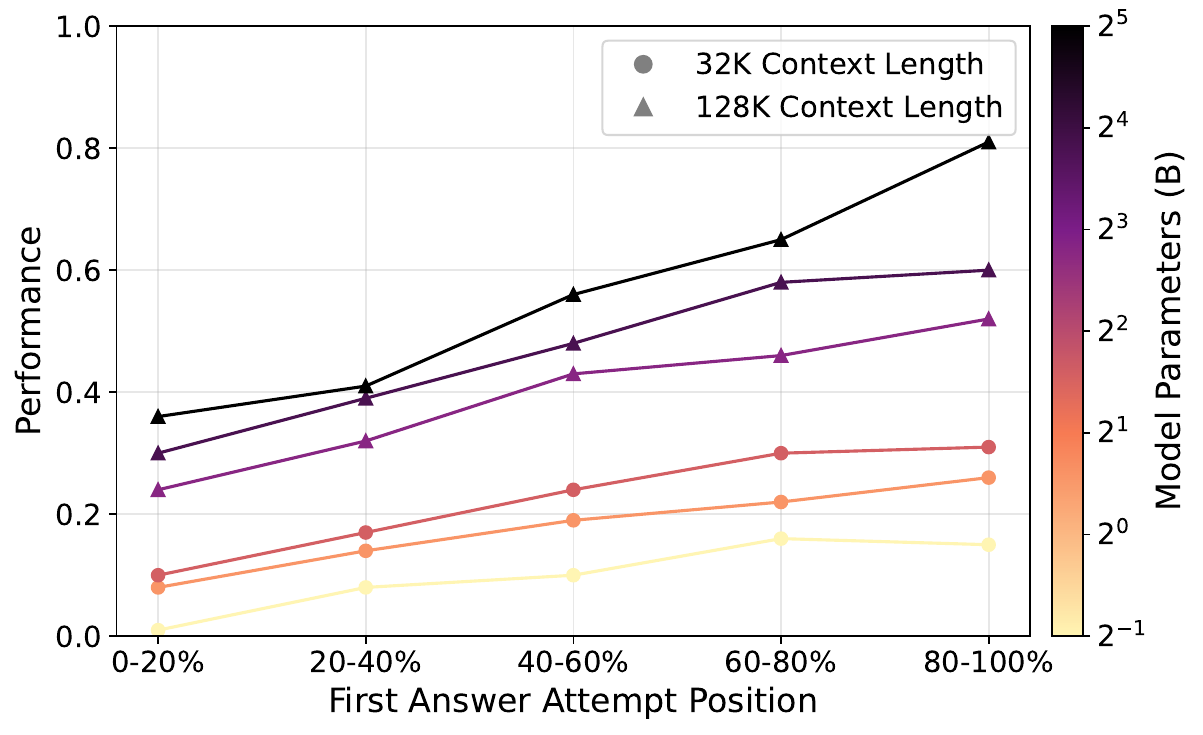}
    \caption{A breakdown of the averaged multi-turn performance of the Qwen2.5-Instruct series, categorized by how early the LLM make its first answer attempt in the conversation.}
    \label{fig:f2}
\end{figure}

To test the effect of RLVR on a model's multi-turn conversation capabilities in simulations of math tasks, we conduct experiments to examine the distribution of its response types before and after training. Furthermore, we analyze the impact of the distribution of its first answer attempt on overall performance. The experimental configuration is described in Sec.~\ref{sec:4.1}. \\
\textbf{Overconfidence in Standard RLVR.}
\label{sec:2.1}
A key challenge in applying RLVR to LLMs stems from the vast solution space of these models contrasted with sparse reward signals, which impedes existing RLVR methods from effectively guiding them toward comprehensive improvements. Because the reward depends entirely on the success or failure of the entire reasoning chain, models are driven toward exploitation of existing capabilities—continuously refining their knowledge and strategies—which constricts exploration and induces gradual capability degradation. As illustrated in Fig.~\ref{fig:f1}, models trained via standard RLVR, when faced with underspecified or ambiguous queries, tend to produce overconfident responses rather than seeking clarification or engaging in further user interaction. \\
\textbf{Why do LLMs need to be humble.}
\label{sec:2.2}
In conversations initiated with under-specified queries, the language model operates with the least amount of information, facing maximum ambiguity. Consequently, the likelihood of formulating a correct solution attempt is at its lowest. Previous studies have demonstrated that an early attempt at a solution can implant flawed assumptions, thereby exerting a misleading influence on subsequent conversational interactions. As illustrated in Fig.~\ref{fig:f2}, we delineate five intervals: if the first answer attempt occurs within the first 20\% of conversation turns, it is categorized into the 0-20\% interval; analogous divisions are applied to the subsequent intervals of 20-40\%, 40-60\%, 60-80\%, and 80-100\%. Our key observation is that for all models, conversations with a later first answer attempt lead to higher performance. Furthermore, a longer context length enhances multi-turn conversation capabilities. This phenomenon thus imposes a key requirement on models: when confronted with ambiguous instructions, they must be humble.

\section{ICPO}
\label{sec:3}
\begin{figure*}
    \centering
    \includegraphics[width=0.75\linewidth]{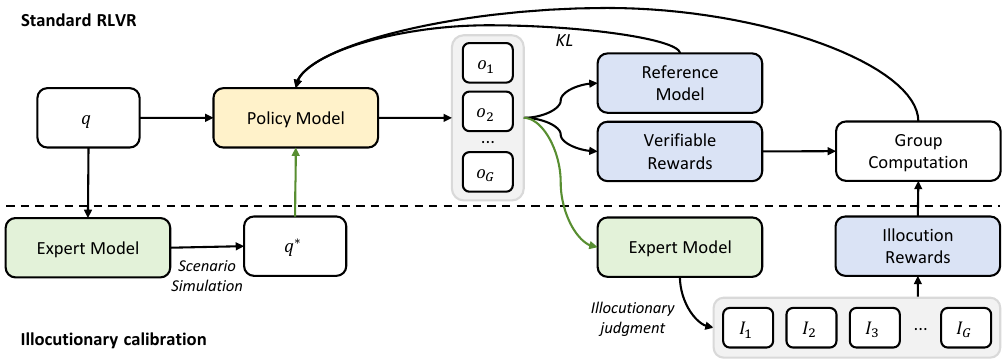}
    \caption{Overview of ICPO. Unlike standard RLVR which rewards only final outcomes, ICPO conditions the reward signal on the user's illocutionary intent. By incorporating underspecified prompts via \textbf{Scenario Simulation} and rewarding calibrated humility through \textbf{Illocutionary Judgment}, the model learns to avoid overconfident hallucinations.}
    \label{fig:f3}
\end{figure*}

To address the aforementioned limitations, we propose ICPO. The core principle of ICPO is to sensitize the model's policy to the illocutionary intent of user instructions, particularly by enabling it to identify and respond to the ambiguity implicit within them. We present the overview of ICPO in Fig.~\ref{fig:f3}.

\subsection{Scenario Simulation}
First, we utilize an expert model $\mathcal{M}_{\text{sim}}$ to process questions $q$, with the objective of transforming it into a situation that might be encountered in real-world conversation. In the context of a conversation scenario involving an underspecified instruction, we prompt the expert model to randomly remove one or two conditions from the question, thereby rendering it an unsolvable question:

\begin{equation}
 q^* = \mathcal{M}_{\text{sim}}(q), 
\end{equation}

\subsection{Illocutionary Judgment}
Subsequently, we employ an expert model $\mathcal{M}_{\text{jud}}$ to determine the response types $\{I_i\}_{i=1}^G$ corresponding to the $G$ candidate responses $\{o_i\}_{i=1}^G$ generated from $q^*$. Motivated by previous studies~\cite{herlihy2024overcomingmiscalibratedconversationalpriors,laban2025llmslostmultiturnconversation}, we similarly define seven response categories: Answer attempt, Clarification, Interrogation, Discussion, Hedging, Refusal, and Missing.

\subsection{Reward Modeling}
For standard questions $q$, we adopt the verifiable reward defined in the standard RLVR framework: 
\begin{equation}
r =
\begin{cases}
1, & \text{if correct}, \\
0, & \text{otherwise}.
\end{cases}
\end{equation}
For simulation scenario questions $q^*$, we define a set of expected response types, denoted as 
\begin{equation}
 \mathcal{T} = \{\text{Clarification}, \text{Interrogation}, \text{Discussion}, \text{Hedging}, \text{Refusal}\},
\end{equation}
If the model's response type belongs to the expected type set $\mathcal{T}$, a positive reward is assigned; otherwise, a zero reward is given:
\begin{equation}
r^* =
\begin{cases}
1, & \text{if the response type } \in \mathcal{T}, \\
0, & \text{otherwise}.
\end{cases}
\end{equation}
The final reward function is:
\begin{equation}
\tilde{r}(o|\tilde{q}) =
\begin{cases}
r(o|\tilde{q}), & \tilde{q}=q, \\
r^*(\mathcal{M}_\text{jud}(o)|\tilde{q}), & \tilde{q}=q^*.
\end{cases}
\end{equation}

\subsection{Reinforcement Learning Algorithms}
We use Group Relative Policy Optimization (GRPO)~\cite{deepseek-math} as our standard RLVR algorithm. Each question $\tilde{q} = \{q, q^*\}$ corresponds to a set of candidate responses $\{o_i\}_{i=1}^G$. We define: 

\begin{equation}
\rho_i = \frac{\pi_{\theta}(o_i \mid \tilde{q})}{\pi_{\theta_{\text{old}}}(o_i \mid \tilde{q})}, \quad A_{i}= \frac{\tilde{r}_i - \text{mean}(\tilde{r})}{\text{std}(\tilde{r})},
\end{equation}
The final loss is then:
\begin{multline}
\mathcal{L}_{\text{ICPO}}(\theta) = \\
\mathbb{E}_{\substack{\tilde{q} \sim \mathcal{D} \\ \{o_i\}\sim \pi_{\theta_{\text{old}}}}} 
\bigg[ \frac{1}{G} \sum_{i=1}^{G} 
\min\!\big( \rho_i A_i, \; \text{clip}(\rho_i, 1-\alpha, 1+\alpha) A_i \big) \\
- \beta D_{\text{KL}}(\pi_\theta \,\|\, \pi_{\text{ref}})\bigg].
\end{multline}

\section{Experiments}
\label{sec:4}
\subsection{Experimental Setups}
\label{sec:4.1}

\textbf{Model.}
We chose the Qwen2.5-Instruct series~\cite{qwen2025qwen25technicalreport} as our backbone, with model sizes ranging from 0.5B to 32B parameters. For scenario simulation and illocutionary judgment, we use DeepSeek-V3-250324~\cite{deepseekai2025deepseekv3technicalreport} as the expert model. \\
\textbf{Evaluation.}
We evaluate the model's multi-turn capability based on the framework from prior research~\cite{laban2025llmslostmultiturnconversation}. This framework simulates real-world scenarios, where a user begins with vague instructions and progressively provides more complete conditions across conversation turns. The evaluation metric is the accuracy of the final response. In this study, we focus on assessing multi-turn performance in mathematical tasks, using a multi-turn conversation dataset adapted from GSM8K~\cite{cobbe2021trainingverifierssolvemath}. In addition, we also evaluate the model's performance on MATH500~\cite{lightman2023lets}, AMC23, AIME24, AIME25, Minerva~\cite{lewkowycz2022solving}, and Olympiad~\cite{he2024olympiadbenchchallengingbenchmarkpromoting}. For the AMC23, AIME24, and AIME25 datasets, we compute the average accuracy over 8 rollouts. During evaluation, we set the rollout temperature to 0.7 and the maximum generation length to 4096. \\
\textbf{Training Details.}
Our codebase is adapted from VeRL~\cite{Sheng_2025}. We trained two models, Qwen2.5-1.5B-Instruct and Qwen2.5-7B-Instruct, on the OpenR1-Math-220k dataset using both standard RVLR and ICPO. Both methods are based on the GRPO~\cite{deepseek-math}. The training is conducted for 600 steps with a batch size of 64, a rollout of 8, and a learning rate of 1e-6. The experiment was conducted on 8 × A800-SXM4-80GB.

\definecolor{mycolorblue}{RGB}{0, 102, 204}
\definecolor{mycolororange}{RGB}{128, 0, 128}  % 新增：橙色，用于Mt.列

\begin{table*}[t]
  \centering
  \caption{Performance of the Qwen2.5-1.5B-Instruct and Qwen2.5-7B-Instruct models on six competition-level math benchmarks, along with average performance (Avg.) and average performance on multi-turn conversations (Mt.). Bold and underline indicate the best and second-best results, respectively.}
  \label{tab:t2}
  \begin{tabular}{lcccccccc}
    \toprule
    \textbf{Model} & \textbf{AIME24} & \textbf{AIME25} & \textbf{AMC23} & \textbf{MATH500} & \textbf{Olympiad} & \textbf{Minerva} & \textbf{Avg.} & \textbf{Mt.} \\
    \midrule
    Qwen2.5-1.5B-Instruct     & 2.5 & 1.2 & 25.3 & 43.8 & 18.1 & 10.6 & \cellcolor{mycolorblue!15}16.9 & \cellcolor{mycolororange!15}17.0 \\
    \quad + Standard RLVR     & 7.8 & 3.7 & 34.4 & 50.4 & 23.9 & 17.0 & \cellcolor{mycolorblue!30}22.9 & \cellcolor{mycolororange!30}26.1 \\
    \quad + \textbf{ICPO (Ours)}       & \textbf{7.9} & \underline{3.2} & \textbf{34.9} & \underline{49.9} & \textbf{24.3} & \textbf{17.7} & \cellcolor{mycolorblue!40}\textbf{23.0} & \cellcolor{mycolororange!45}\textbf{32.8} \\
    \midrule
    Qwen2.5-7B-Instruct       & 11.7 & 6.6 & 36.4 & 67.3 & 29.4 & 14.5 & \cellcolor{mycolorblue!47}27.7 & \cellcolor{mycolororange!30}35.4 \\
    \quad + Standard RLVR     & 13.3 & 7.9 & 49.8 & 80.7 & 43.3 & 31.5 & \cellcolor{mycolorblue!60}37.8 & \cellcolor{mycolororange!45}47.2 \\ 
    \quad + \textbf{ICPO (Ours)}      & \underline{12.9} & \underline{7.7} & \textbf{51.9} & \textbf{81.2} & \textbf{44.8} & \underline{30.4} & \cellcolor{mycolorblue!77}\textbf{38.2} & \cellcolor{mycolororange!80}\textbf{55.4} \\  % 100%为纯色
    \bottomrule
  \end{tabular}
\end{table*}

% --- 新增表格开始 ---
\begin{table}[t]
  \centering
  \caption{Comparison of Multi-turn (Mt.) performance on Qwen2.5-7B-Instruct against various entropy-regularization and anti-collapse RLVR variants.}
  \label{tab:baselines}
  \begin{tabular}{lc}
    \toprule
    \textbf{Method} & \textbf{Mt.} \\
    \midrule
    Qwen2.5-7B-Instruct      & 35.4 \\
    GRPO                     & 47.2 \\
    \quad + Clip-higher      & 48.9 \\
    \quad + Clip-Cov         & 50.2 \\
    \quad + KL-Cov           & 47.7 \\
    RL-PLUS                  & 51.6 \\
    \textbf{ICPO (Ours)}     & \textbf{55.4} \\
    \bottomrule
  \end{tabular}
\end{table}
% --- 新增表格结束 ---

\begin{figure}[t]
    \centering
    \includegraphics[width=0.85\linewidth]{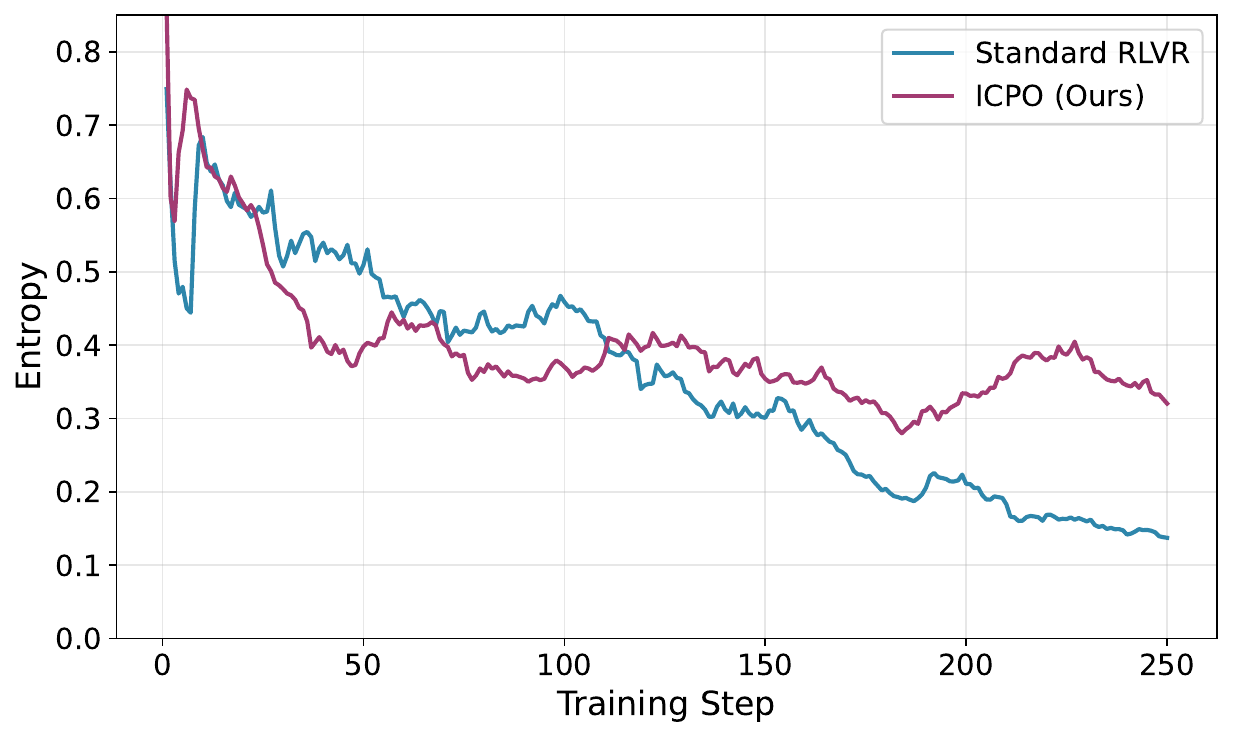}
    \caption{Entropy comparison between Standard RLVR and ICPO during training.}
    \label{fig:entropy}
\end{figure}

\subsection{Main Results}
$\bullet$ \textbf{Greater Humility.} 
Fig.~\ref{fig:f1} provides a compelling illustration of the dynamic shifts in response types following model training. To ensure the analysis focuses purely on model behavior rather than technical limitations, we have excluded statistics for the ``Missing'' type, which can arise from context length constraints. The results clearly indicate that after ICPO training, the model adopts a more humble and cautious stance when confronted with ambiguous instructions. This is evidenced by a remarkable 93\% increase in the proportion of other response types (e.g., Clarification, Refusal). In contrast, the standard RLVR approach tends to amplify the model's propensity for direct answers, potentially increasing the risk of overconfident and incorrect responses. \\
$\bullet$ \textbf{Better multi-turn conversation.} 
As detailed in Tab.~\ref{tab:t2}, we present a comparative analysis of model performance in both mathematical reasoning and multi-turn conversation following training with standard RLVR and our proposed ICPO method, respectively. We observe that the standard RLVR model exhibits a modest gain in multi-turn accuracy, an improvement we attribute to its enhanced mathematical capabilities. In contrast, our ICPO method not only achieves a slight competitive edge on mathematical tasks but, more critically, delivers a substantial enhancement to the model's multi-turn conversational capabilities, boasting an average improvement of approximately 75\% over the RLVR baseline. \\
$\bullet$ \textbf{Higher Response Diversity.} 
As illustrated in Fig.~\ref{fig:entropy}, our proposed method fosters greater response diversity by maintaining a stable and high policy entropy, fluctuating around 0.3. This elevated entropy signifies a flatter, more varied probability distribution over the action space, which prevents the model from prematurely converging on suboptimal solutions. In stark contrast, the standard RLVR baseline suffers from mode collapse as its entropy rapidly decreases to a near-deterministic level of 0.1. This indicates an overconfident policy that has settled on a narrow set of outputs. By avoiding this, our approach preserves greater diversity in its generations and maintains the flexibility to produce more nuanced and varied responses. \\
$\bullet$ \textbf{Superiority over Entropy Regularizers.}
To verify that ICPO's effectiveness stems from specific illocutionary calibration rather than merely preventing policy collapse, we conducted controlled experiments comparing ICPO against other RLVR variants equipped with entropy-regularization or anti-collapse mechanisms, including GRPO with Clip-higher \cite{yu2025dapoopensourcellmreinforcement}, Clip-Cov, KL-Cov \cite{cui2025entropymechanismreinforcementlearning}, and RL-PLUS \cite{dong2025rlpluscounteringcapabilityboundary}. As shown in Tab.~\ref{tab:baselines}, while generic regularization methods alleviate the ``lost-in-conversation'' issue to some extent, ICPO achieves consistently larger improvements (55.4\%). This demonstrates that conditioning rewards on illocutionary intent provides a more precise guidance for maintaining multi-turn robustness than indiscriminate entropy maintenance. \\

\begin{figure}[t]
    \centering
    \includegraphics[width=0.90\linewidth]{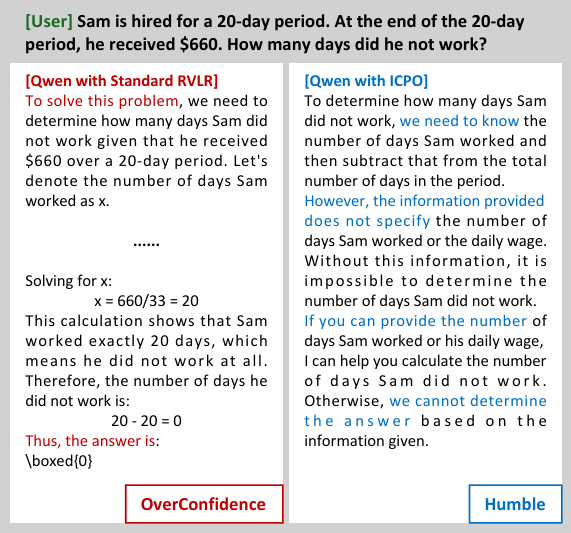}
    \caption{Conversation example, in which the user starts the conversation with a problem that has insufficient conditions and cannot be solved.}
    \label{fig:f4}
\end{figure}

\subsection{Case Study}
Fig.~\ref{fig:f4} illustrates a real-world scenario initiated by an under-specified user instruction. The problem as stated is inherently unsolvable due to the absence of critical conditions, such as the value of ``daily rewards''. When presented with this ambiguity, the baseline model trained with standard RLVR produces a factually incorrect, overconfident, and hallucinatory response. In stark contrast, the model trained with our proposed ICPO method demonstrates a more robust and cautious behavior. It correctly analyzes the prompt to identify the missing prerequisites, proactively requests the necessary information from the user, and candidly admits its inability to determine a definitive answer without further details.

\section{Conclusion}
In this paper, we identify a critical limitation of standard RLVR: it inadvertently promotes overconfidence, causing LLMs to struggle with ambiguous multi-turn conversations. To address this, we propose ICPO, a novel framework that sensitizes the model to instruction ambiguity. By dynamically rewarding the model for seeking clarification rather than forcing direct answers, ICPO achieves a significant 75\% improvement in multi-turn reasoning while preserving robust single-turn performance. Crucially, our comparative experiments demonstrate that ICPO consistently outperforms generic entropy-regularization and anti-collapse mechanisms. This finding underscores that maintaining conversational robustness requires targeted illocutionary guidance rather than merely enforcing policy diversity. Our work offers a practical path toward more collaborative AI agents. Future work will focus on scaling ICPO to larger models and verifying its efficacy across broader, non-mathematical conversational contexts.

% \section{Funding Acknowledgements}
% The authors would like to acknowledge the support of the National Natural Science Foundation of China under Grant No. 62376246.

\bibliographystyle{IEEEbib}
\bibliography{refs}

@inproceedings{herlihy2024overcomingmiscalibratedconversationalpriors,
author = {Herlihy, Christine and Neville, Jennifer and Schnabel, Tobias and Swaminathan, Adith},
title = {On overcoming miscalibrated conversational priors in LLM-based chatbots},
year = {2024},
publisher = {JMLR.org},
booktitle = {Proceedings of the Fortieth Conference on Uncertainty in Artificial Intelligence},
articleno = {75},
numpages = {22},
location = {Barcelona, Spain},
series = {UAI '24}
}

@misc{laban2025llmslostmultiturnconversation,
      title={LLMs Get Lost In Multi-Turn Conversation}, 
      author={Philippe Laban and Hiroaki Hayashi and Yingbo Zhou and Jennifer Neville},
      year={2025},
      eprint={2505.06120},
      archivePrefix={arXiv},
      primaryClass={cs.CL},
      url={https://arxiv.org/abs/2505.06120}, 
}

@misc{cui2025entropymechanismreinforcementlearning,
      title={The Entropy Mechanism of Reinforcement Learning for Reasoning Language Models}, 
      author={Ganqu Cui and Yuchen Zhang and Jiacheng Chen and Lifan Yuan and Zhi Wang and Yuxin Zuo and Haozhan Li and Yuchen Fan and Huayu Chen and Weize Chen and Zhiyuan Liu and Hao Peng and Lei Bai and Wanli Ouyang and Yu Cheng and Bowen Zhou and Ning Ding},
      year={2025},
      eprint={2505.22617},
      archivePrefix={arXiv},
      primaryClass={cs.LG},
      url={https://arxiv.org/abs/2505.22617}, 
}

@misc{deepseek-math,
  author = {Zhihong Shao and Peiyi Wang and Qihao Zhu and Runxin Xu and Junxiao Song and Mingchuan Zhang and Y.K. Li and Y. Wu and Daya Guo},
  title = {DeepSeekMath: Pushing the Limits of Mathematical Reasoning in Open Language Models},
  journal = {CoRR},
  volume = {abs/2402.03300},
  year = {2024},
  url = {https://arxiv.org/abs/2402.03300},
}

@misc{deepseekai2025deepseekv3technicalreport,
      title={DeepSeek-V3 Technical Report}, 
      author={DeepSeek-AI and Aixin Liu and Bei Feng and Bing Xue and Bingxuan Wang and Bochao Wu and Chengda Lu and Chenggang Zhao and Chengqi Deng and Chenyu Zhang and Chong Ruan and Damai Dai and Daya Guo and Dejian Yang and Deli Chen and Dongjie Ji and Erhang Li and Fangyun Lin and others},
      year={2025},
      eprint={2412.19437},
      archivePrefix={arXiv},
      primaryClass={cs.CL},
      url={https://arxiv.org/abs/2412.19437}, 
}

@inproceedings{Sheng_2025, series={EuroSys ’25},
   title={HybridFlow: A Flexible and Efficient RLHF Framework},
   url={http://dx.doi.org/10.1145/3689031.3696075},
   DOI={10.1145/3689031.3696075},
   booktitle={Proceedings of the Twentieth European Conference on Computer Systems},
   publisher={ACM},
   author={Sheng, Guangming and Zhang, Chi and Ye, Zilingfeng and Wu, Xibin and Zhang, Wang and Zhang, Ru and Peng, Yanghua and Lin, Haibin and Wu, Chuan},
   year={2025},
   month=mar, pages={1279–1297},
   collection={EuroSys ’25} }

@misc{qwen2025qwen25technicalreport,
      title={Qwen2.5 Technical Report}, 
      author={An Yang and Baosong Yang and Beichen Zhang and Binyuan Hui and Bo Zheng and Bowen Yu and Chengyuan Li and Dayiheng Liu and Fei Huang and Haoran Wei and Huan Lin and Jian Yang and Jianhong Tu and Jianwei Zhang and Jianxin Yang and Jiaxi Yang and others},
      year={2025},
      eprint={2412.15115},
      archivePrefix={arXiv},
      primaryClass={cs.CL},
      url={https://arxiv.org/abs/2412.15115}, 
}

@article{kong2025survey,
  title={A Survey of LLM-Driven AI Agent Communication: Protocols, Security Risks, and Defense Countermeasures},
  author={Kong, Dezhang and Lin, Shi and Xu, Zhenhua and Wang, Zhebo and Li, Minghao and Li, Yufeng and Zhang, Yilun and Peng, Hujin and Sha, Zeyang and Li, Yuyuan and others},
  journal={arXiv preprint arXiv:2506.19676},
  year={2025}
}

@inproceedings{liu2025understandingr1zeroliketrainingcritical,
  title={Understanding r1-zero-like training: A critical perspective},
  author={Liu, Zichen and Chen, Changyu and Li, Wenjun and Qi, Penghui and Pang, Tianyu and Du, Chao and Lee, Wee Sun and Lin, Min},
  booktitle={Conference on Language Modeling (COLM)},
  year={2025}
}

@misc{dong2025rlpluscounteringcapabilityboundary,
      title={RL-PLUS: Countering Capability Boundary Collapse of LLMs in Reinforcement Learning with Hybrid-policy Optimization}, 
      author={Yihong Dong and Xue Jiang and Yongding Tao and Huanyu Liu and Kechi Zhang and Lili Mou and Rongyu Cao and Yingwei Ma and Jue Chen and Binhua Li and Zhi Jin and Fei Huang and Yongbin Li and Ge Li},
      year={2025},
      eprint={2508.00222},
      archivePrefix={arXiv},
      primaryClass={cs.AI},
      url={https://arxiv.org/abs/2508.00222}, 
}

@inproceedings{
yu2025dapoopensourcellmreinforcement,
title={{DAPO}: An Open-Source {LLM} Reinforcement Learning System at Scale},
author={Qiying Yu and Zheng Zhang and Ruofei Zhu and Yufeng Yuan and Xiaochen Zuo and YuYue and Weinan Dai and Tiantian Fan and Gaohong Liu and Juncai Liu and LingJun Liu and Xin Liu and Haibin Lin and Zhiqi Lin and Bole Ma and Guangming Sheng and Yuxuan Tong and Chi Zhang and Mofan Zhang and Ru Zhang and Wang Zhang and Hang Zhu and Jinhua Zhu and Jiaze Chen and Jiangjie Chen and Chengyi Wang and Hongli Yu and Yuxuan Song and Xiangpeng Wei and Hao Zhou and Jingjing Liu and Wei-Ying Ma and Ya-Qin Zhang and Lin Yan and Yonghui Wu and Mingxuan Wang},
booktitle={The Thirty-ninth Annual Conference on Neural Information Processing Systems},
year={2025},
url={https://openreview.net/forum?id=2a36EMSSTp}
}

@misc{zhang2025surveyreinforcementlearninglarge,
      title={A Survey of Reinforcement Learning for Large Reasoning Models}, 
      author={Kaiyan Zhang and Yuxin Zuo and Bingxiang He and Youbang Sun and Runze Liu and Che Jiang and Yuchen Fan and Kai Tian and Guoli Jia and Pengfei Li and Yu Fu and Xingtai Lv and Yuchen Zhang and Sihang Zeng and Shang Qu and Haozhan Li and Shijie Wang and Yuru Wang and Xinwei Long and Fangfu Liu and Xiang Xu and Jiaze Ma and Xuekai Zhu and Ermo Hua and Yihao Liu and Zonglin Li and Huayu Chen and Xiaoye Qu and Yafu Li and Weize Chen and Zhenzhao Yuan and Junqi Gao and Dong Li and Zhiyuan Ma and Ganqu Cui and Zhiyuan Liu and Biqing Qi and Ning Ding and Bowen Zhou},
      year={2025},
      eprint={2509.08827},
      archivePrefix={arXiv},
      primaryClass={cs.CL},
      url={https://arxiv.org/abs/2509.08827}, 
}

@inproceedings{
lightman2023lets,
title={Let's Verify Step by Step},
author={Hunter Lightman and Vineet Kosaraju and Yuri Burda and Harrison Edwards and Bowen Baker and Teddy Lee and Jan Leike and John Schulman and Ilya Sutskever and Karl Cobbe},
booktitle={The Twelfth International Conference on Learning Representations},
year={2024},
url={https://openreview.net/forum?id=v8L0pN6EOi}
}

@misc{cobbe2021trainingverifierssolvemath,
      title={Training Verifiers to Solve Math Word Problems}, 
      author={Karl Cobbe and Vineet Kosaraju and Mohammad Bavarian and Mark Chen and Heewoo Jun and Lukasz Kaiser and Matthias Plappert and Jerry Tworek and Jacob Hilton and Reiichiro Nakano and Christopher Hesse and John Schulman},
      year={2021},
      eprint={2110.14168},
      archivePrefix={arXiv},
      primaryClass={cs.LG},
      url={https://arxiv.org/abs/2110.14168}, 
}

@inproceedings{he2024olympiadbenchchallengingbenchmarkpromoting,
    title = "{O}lympiad{B}ench: A Challenging Benchmark for Promoting {AGI} with Olympiad-Level Bilingual Multimodal Scientific Problems",
    author = "He, Chaoqun  and
      Luo, Renjie  and
      Bai, Yuzhuo  and
      Hu, Shengding  and
      Thai, Zhen  and
      Shen, Junhao  and
      Hu, Jinyi  and
      Han, Xu  and
      Huang, Yujie  and
      Zhang, Yuxiang  and
      Liu, Jie  and
      Qi, Lei  and
      Liu, Zhiyuan  and
      Sun, Maosong",
    editor = "Ku, Lun-Wei  and
      Martins, Andre  and
      Srikumar, Vivek",
    booktitle = "Proceedings of the 62nd Annual Meeting of the Association for Computational Linguistics (Volume 1: Long Papers)",
    month = aug,
    year = "2024",
    address = "Bangkok, Thailand",
    publisher = "Association for Computational Linguistics",
    url = "https://aclanthology.org/2024.acl-long.211/",
    doi = "10.18653/v1/2024.acl-long.211",
    pages = "3828--3850"
}

@inproceedings{lewkowycz2022solving,
  title        = {Solving quantitative reasoning problems with language models},
  author       = {Lewkowycz, Aitor and Andreassen, Anders and Dohan, David and Dyer, Ethan and Michalewski, Henryk and Ramasesh, Vinay and Slone, Ambrose and Anil, Cem and Schlag, Imanol and Gutman-Solo, Theo and others},
  booktitle    = {Advances in Neural Information Processing Systems},
  volume       = {35},
  pages        = {3843--3857},
  year         = {2022}
}

@misc{xu2025copyrightprotectionlargelanguage,
      title={Copyright Protection for Large Language Models: A Survey of Methods, Challenges, and Trends}, 
      author={Zhenhua Xu and Xubin Yue and Zhebo Wang and Qichen Liu and Xixiang Zhao and Jingxuan Zhang and Wenjun Zeng and Wengpeng Xing and Dezhang Kong and Changting Lin and Meng Han},
      year={2025},
      eprint={2508.11548},
      archivePrefix={arXiv},
      primaryClass={cs.CR},
      url={https://arxiv.org/abs/2508.11548}, 
}

@misc{r1,
      title={SRAF: Stealthy and Robust Adversarial Fingerprint for Copyright Verification of Large Language Models}, 
      author={Zhebo Wang and Zhenhua Xu and Maike Li and Wenpeng Xing and Chunqiang Hu and Chen Zhi and Meng Han},
      year={2026},
      eprint={2505.06304},
      archivePrefix={arXiv},
      primaryClass={cs.CR},
      url={https://arxiv.org/abs/2505.06304}, 
}

@misc{r2,
      title={Fingerprint Vector: Enabling Scalable and Efficient Model Fingerprint Transfer via Vector Addition}, 
      author={Zhenhua Xu and Qichen Liu and Zhebo Wang and Wenpeng Xing and Dezhang Kong and Mohan Li and Meng Han},
      year={2025},
      eprint={2409.08846},
      archivePrefix={arXiv},
      primaryClass={cs.CR},
      url={https://arxiv.org/abs/2409.08846}, 
}

@misc{r3,
      title={Spectral Logit Sculpting: Adaptive Low-Rank Logit Transformation for Controlled Text Generation}, 
      author={Jin Li and Zhebo Wang and Tianliang Lu and Mohan Li and Wenpeng Xing and Meng Han},
      year={2025},
      eprint={2509.25204},
      archivePrefix={arXiv},
      primaryClass={cs.LG},
      url={https://arxiv.org/abs/2509.25204}, 
}

@misc{r4,
      title={MEraser: An Effective Fingerprint Erasure Approach for Large Language Models}, 
      author={Jingxuan Zhang and Zhenhua Xu and Rui Hu and Wenpeng Xing and Xuhong Zhang and Meng Han},
      year={2025},
      eprint={2506.12551},
      archivePrefix={arXiv},
      primaryClass={cs.CR},
      url={https://arxiv.org/abs/2506.12551}, 
}

@misc{r5,
      title={ForgetMark: Stealthy Fingerprint Embedding via Targeted Unlearning in Language Models}, 
      author={Zhenhua Xu and Haobo Zhang and Zhebo Wang and Qichen Liu and Haitao Xu and Wenpeng Xing and Meng Han},
      year={2026},
      eprint={2601.08189},
      archivePrefix={arXiv},
      primaryClass={cs.CR},
      url={https://arxiv.org/abs/2601.08189}, 
}

@inproceedings{r6,
    title = "{CTCC}: A Robust and Stealthy Fingerprinting Framework for Large Language Models via Cross-Turn Contextual Correlation Backdoor",
    author = "Xu, Zhenhua  and
      Zhao, Xixiang  and
      Yue, Xubin  and
      Tian, Shengwei  and
      Lin, Changting  and
      Han, Meng",
    editor = "Christodoulopoulos, Christos  and
      Chakraborty, Tanmoy  and
      Rose, Carolyn  and
      Peng, Violet",
    booktitle = "Proceedings of the 2025 Conference on Empirical Methods in Natural Language Processing",
    month = nov,
    year = "2025",
    address = "Suzhou, China",
    publisher = "Association for Computational Linguistics",
    url = "https://aclanthology.org/2025.emnlp-main.356/",
    doi = "10.18653/v1/2025.emnlp-main.356",
    pages = "6967--6989",
    ISBN = "979-8-89176-332-6"
}

@inproceedings{r7,
    title = "{E}ver{T}racer: Hunting Stolen Large Language Models via Stealthy and Robust Probabilistic Fingerprint",
    author = "Xu, Zhenhua  and
      Han, Meng  and
      Xing, Wenpeng",
    editor = "Christodoulopoulos, Christos  and
      Chakraborty, Tanmoy  and
      Rose, Carolyn  and
      Peng, Violet",
    booktitle = "Proceedings of the 2025 Conference on Empirical Methods in Natural Language Processing",
    month = nov,
    year = "2025",
    address = "Suzhou, China",
    publisher = "Association for Computational Linguistics",
    url = "https://aclanthology.org/2025.emnlp-main.358/",
    doi = "10.18653/v1/2025.emnlp-main.358",
    pages = "7008--7031",
    ISBN = "979-8-89176-332-6"
}

@misc{r8,
      title={AdaMARP: An Adaptive Multi-Agent Interaction Framework for General Immersive Role-Playing},
      author={Zhenhua Xu and Dongsheng Chen and Shuo Wang and Jian Li and Chengjie Wang and Meng Han and Yabiao Wang},
      year={2026},
      eprint={2601.11007},
      archivePrefix={arXiv},
      primaryClass={cs.AI},
      url={https://arxiv.org/abs/2601.11007}, 
}

@inproceedings{r9,
    title = "Unlocking the Effectiveness of {L}o{RA}-{FP} for Seamless Transfer Implantation of Fingerprints in Downstream Models",
    author = "Xu, Zhenhua  and
      Yan, Zhaokun  and
      Xu, Binhan  and
      Tong, Xin  and
      Xu, Haitao  and
      Chen, Yourong  and
      Han, Meng",
    editor = "Christodoulopoulos, Christos  and
      Chakraborty, Tanmoy  and
      Rose, Carolyn  and
      Peng, Violet",
    booktitle = "Findings of the Association for Computational Linguistics: EMNLP 2025",
    month = nov,
    year = "2025",
    address = "Suzhou, China",
    publisher = "Association for Computational Linguistics",
    url = "https://aclanthology.org/2025.findings-emnlp.230/",
    doi = "10.18653/v1/2025.findings-emnlp.230",
    pages = "4302--4312",
    ISBN = "979-8-89176-335-7"
}

@inproceedings{r10,
    title = "{PREE}: Towards Harmless and Adaptive Fingerprint Editing in Large Language Models via Knowledge Prefix Enhancement",
    author = "Yue, Xubin  and
      Xu, Zhenhua  and
      Xing, Wenpeng  and
      Yu, Jiahui  and
      Li, Mohan  and
      Han, Meng",
    editor = "Christodoulopoulos, Christos  and
      Chakraborty, Tanmoy  and
      Rose, Carolyn  and
      Peng, Violet",
    booktitle = "Findings of the Association for Computational Linguistics: EMNLP 2025",
    month = nov,
    year = "2025",
    address = "Suzhou, China",
    publisher = "Association for Computational Linguistics",
    url = "https://aclanthology.org/2025.findings-emnlp.204/",
    doi = "10.18653/v1/2025.findings-emnlp.204",
    pages = "3794--3804",
    ISBN = "979-8-89176-335-7"
}

\end{document}